\documentclass[conference]{IEEEtran}
\IEEEoverridecommandlockouts
\usepackage{cite}
\usepackage{amsmath,amssymb,amsfonts}
\usepackage{algorithmic}
\usepackage{graphicx}
\usepackage{textcomp}
\usepackage{xcolor}

\def\BibTeX{{\rm B\kern-.05em{\sc i\kern-.025em b}\kern-.08em T\kern-.1667em\lower.7ex\hbox{E}\kern-.125emX}}

\begin{document}

\title{Explainable AI based Glaucoma Detection using Transfer Learning and LIME\\
}

\author{\IEEEauthorblockN{Touhidul Islam Chayan}
\IEEEauthorblockA{\textit{Computer Science and Engineering} \\
\textit{BRAC University}\\
Dhaka, Bangladesh \\
touhidul.islam.chayan@g.bracu.ac.bd}
\and
\IEEEauthorblockN{Anita Islam}
\IEEEauthorblockA{\textit{Computer Science and Engineering} \\
\textit{BRAC University}\\
Dhaka, Bangladesh \\
anita.islam@g.bracu.ac.bd}
\and
\IEEEauthorblockN{Eftykhar Rahman}
\IEEEauthorblockA{\textit{Computer Science and Engineering} \\
\textit{BRAC University}\\
Dhaka, Bangladesh \\
eftykhar.rahman@g.bracu.ac.bd}
\and
\IEEEauthorblockN{Md. Tanzim Reza}
\IEEEauthorblockA{\textit{Computer Science and Engineering} \\
\textit{BRAC University}\\
Dhaka, Bangladesh \\
tanzim.reza@bracu.ac.bd}
\and
\IEEEauthorblockN{Tasnim Sakib Apon}
\IEEEauthorblockA{\textit{Computer Science and Engineering} \\
\textit{BRAC University}\\
Dhaka, Bangladesh \\
sakibapon7@gmail.com}
\and
\IEEEauthorblockN{MD. Golam Rabiul Alam}
\IEEEauthorblockA{\textit{Computer Science and Engineering} \\
\textit{BRAC University}\\
Dhaka, Bangladesh \\
rabiul.alam@bracu.ac.bd}
}

\maketitle

\begin{abstract}
    Glaucoma is the second driving reason for partial or complete blindness among all the visual deficiencies which mainly occurs because of excessive pressure in the eye due to anxiety or depression which damages the optic nerve and creates complications in vision. Traditional glaucoma screening is a time-consuming process that necessitates the medical professionals' constant attention, and even so time to time due to the time constrains and pressure they fail to classify correctly that leads to wrong treatment. Numerous efforts have been made to automate the entire glaucoma classification procedure however, these existing models in general have a black box characteristics that prevents users from understanding the key reasons behind the prediction and thus medical practitioners generally can not rely on these system. In this article after comparing with various pre-trained models, we propose a transfer learning model that is able to classify Glaucoma with 94.71\% accuracy. In addition, we have utilized Local Interpretable Model-Agnostic Explanations(LIME) that introduces explainability in our system. This improvement enables medical professionals obtain important and comprehensive information that aid them in making judgments. It also lessen the opacity and fragility of the traditional deep learning models.
\end{abstract}


\section{Introduction}
Glaucoma is a very common group of eye diseases caused by damage to the optic nerve that connects the eye to the brain and if untreated, it causes permanent loss of vision which is the second most popular cause of blindness globally. It is also known as the ‘silent thief of sight’ as it cannot be detected at a very early stage \cite{Salam}. Around 57.5 million people worldwide are affected by Glaucoma \cite{Allison}. There are two significant kinds of glaucoma: open-angle and angle-closure. Movement of glaucoma can be halted with medicines, however, part of the vision that is now lost can’t be reestablished. This is the reason it’s vital to distinguish early indications of glaucoma with standard eye tests. Acute angle-closure glaucoma is a visual crisis and requires quick consideration through early detection. Glaucoma can be diagnosed and partial or complete blindness could be prevented if we can detect it in an early stage. \par
Unfortunately, not many people bother about the early detection of Glaucoma whereas it can be diagnosed early to prevent eyesight loss. For this reason, we have decided to work with the early detection of glaucoma disease and going to use Explainable AI (XAI) to classify scanned images of eyes that have glaucoma that proposes the report to the decision of Artificial Intelligence which means Deep Learning or Black Box to the extent that is human interpretable. Moreover, we intend to give an outline of ongoing distributions in regards to the utilization of man-made consciousness to improve the recognition and treatment of glaucoma. Deep Learning (DL) is a subset of Artificial Intelligence (AI) dependent on profound neural networks which have made striking leaps forwards in clinical imaging, especially for image characterization and pattern acknowledgement \cite{Ran}. The main purpose of this study is to represent whether and how deep learning based measurements can be utilized for glaucoma execution in the clinic \cite{Aleci}. On the other hand, if the vision loss has already occurred, the treatment can delay or hinder further vision loss \cite{Diabetic}. Open-angle glaucoma is the most common form of glaucoma and is responsible for 90\% of the cases \cite{Types}. Fundus pictures can be utilized for glaucoma finding through the CDR strategy \cite{aba}. Such CNN models can work in pairs with human specialists to keep up with large eye health and assist recognition of visual deficiency causing eye sickness \cite{Thakoor}. Our objective are as follows: (i) Automating the process for Glaucoma categorization. (ii) Provide detailed information to medical professionals against a prediction, so that they can rely on the system. (iii) Increase the efficiency of the entire process. (iv) Reduce the amount of work required by the medical personnel. Additionally, we encountered a number of challenges or obstacles while performing our study, including dealing with the tendency of models to overfit data, computing resources, etc. \par
The significant contributions of this article are stated as follows:
\begin{itemize}

    \item A transfer learning based Glaucoma disease detection model is proposed and a comprehensive study between various pre-trained model's performance on Glaucoma is conducted. 
    \item Evaluating the interpretability of the proposed model using Local Interpretable Model-Agnostic Explanations(LIME) that offeres the medical practitioners with key features or information for the accurate classification of visual diseased Glaucoma.
    \item Performance of the proposed model has been studied on a benchmark Glaucoma dataset.  
\end{itemize}

A brief overview of previous Glaucoma disease classification research is included in Section II of this paper, followed by a brief discussion of our methodology, models, and techniques in Section III, which is divided into five sub-sections. The explanation of the work plan  is provided in section III-A, whereas III-B discusses data set description, and Section III-C exposes our proposed CNN model. The performance evaluation have been depicted in Section IV. Finally, in Section V we have interpreted our model and attempted to illustrate how it makes decisions.

\section{Literature Review}

    Glaucoma is one of the most common causes of permanent blindness around the world \cite{Lim}. As when the pressure inside the eye is too high in a particular nerve that moment glaucoma will develop and it will also create eye ache. The working mechanisms of the different diagnosis tools like tonometers, gonioscopy, scanning laser tomography, etc are available for the treatment and detection but there are some advantages and disadvantages which sometimes create boundaries. For this, there should be an evaluation of how this works. But with using deep learning the boundaries can be removed. As the XAI concept can be understood by humans which will be closer to the human brain to understand. We have utilized ImageNet's various pre-trained models in order to classify diseased Glaucoma. \par

    \begin{table}[!t]
        \centering
        \caption{Comparison Between the previous studies}
        \begin{tabular}{|c|c|c|c|}
                \hline
                Architectures   & Dataset   & Accuracy   & Reference    \\ \hline
                CNN             & Retinal OCT & 94.87\%  & \cite{apon}  \\ \hline
                ResNet50-v1     & Retinal OCT & 94.92 \% & \cite{Zhang} \\ \hline
                CNN             & Glaucoma    & 84.50\%  & \cite{Abbas} \\ \hline
                FNN         & Glaucoma    & 92.5\%   & \cite{Dervisevic}\\ \hline
        \end{tabular}
        \label{RelatedWorkTable}
    \end{table}
    
    Table \ref{RelatedWorkTable} depicts the brief illustration of previous studies. Additionally, one more research was done from which We learned The impact of artificial intelligence in the diagnosis and management of glaucoma from \cite{Mayro}. Computerized automated visual field testing represents a significant improvement in mapping the island of vision, allowing visual field testing to become a cornerstone in diagnosing and managing glaucoma. Goldbaum developed a two-layer neural network for analyzing visual fields in 1994 et al.\cite{aba}. This network classified normal and glaucomatous eyes with the same sensitivity (65\%) and specificity (72\%) as two glaucoma specialists. \par
    The pathogenesis of glaucoma appears to be dependent on several interconnected pathogenetic mechanisms, including mechanical effects characterized by excessive intraocular pressure, reduced neutrophil produce, hypoxia, excitotoxicity, oxidative stress, and the involvement of autoimmune processes, according to new evidence \cite{Greco}. Hearing loss has also been linked to the development of glaucoma. In normal tension glaucoma patients with hearing loss, antiphosphatidylserine antibodies of the immunoglobulin G class were shown to be more prevalent than in normal-tension glaucoma patients with normacusis. The World Health Organization reports that glaucoma affects approximately 60 million people worldwide. By the year 2020, it is expected that approximately 80 million people will suffer from glaucoma, which is anticipated to result in 11.2 million cases of bilateral blindness \cite{Pascolini}. This is why it needs to be treated as early as possible according to the authors. \par
    Unlike the studies mentioned above, our focus has been on interpreting our proposed model such that medical practitioners would feel confident utilizing our approach.
\section{Methodology}
    We can obtain a clear overview of our proposed model which is separated into three subsections, from Section III. Part III-A discusses about our working plan, followed by part III-B, which discusses data gathering and pre-processing, and lastly, part III-C, which discusses the architecture.
    \subsection{System Model}
    
        \begin{figure*}[!t]
            \centerline{\includegraphics[scale=0.1]{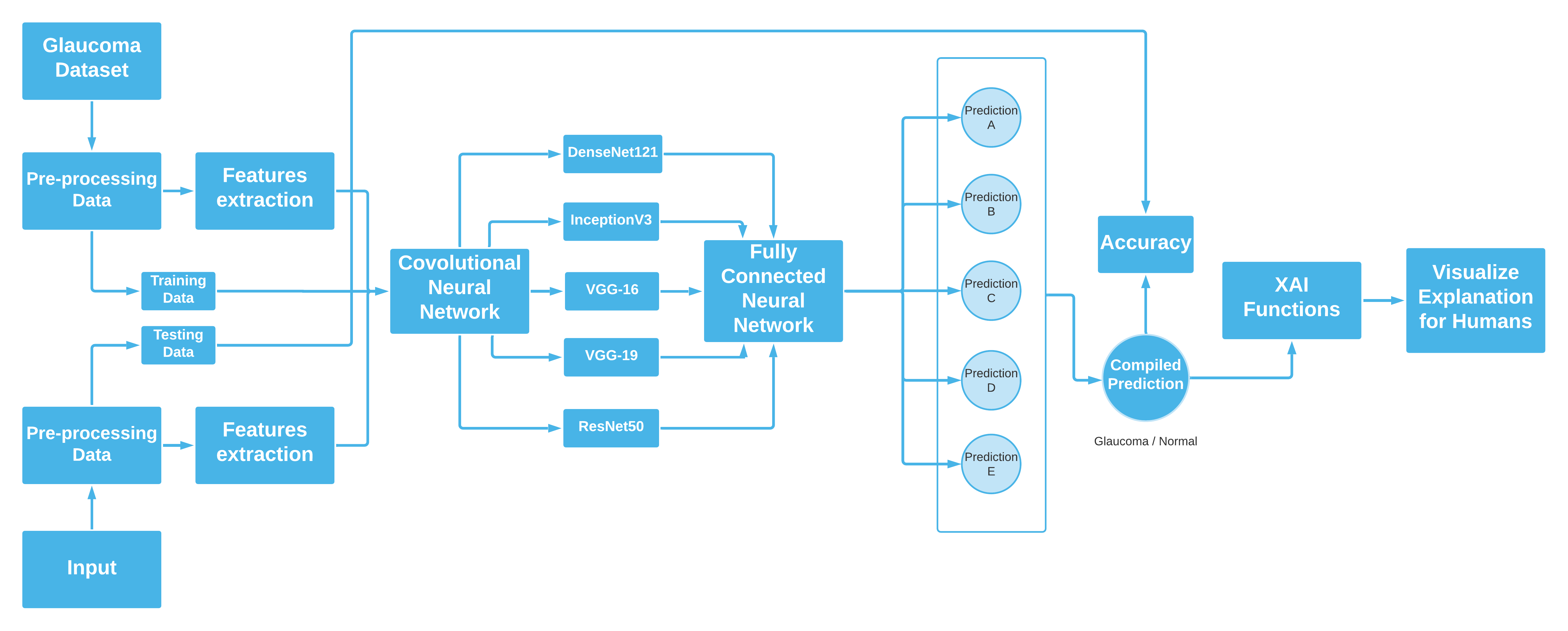}}
            \caption{System Model: Glaucoma Classification}
            \label{SystemModel}
        \end{figure*}
    
\begin{figure}[!t]
        \centerline{\includegraphics[scale=0.5]{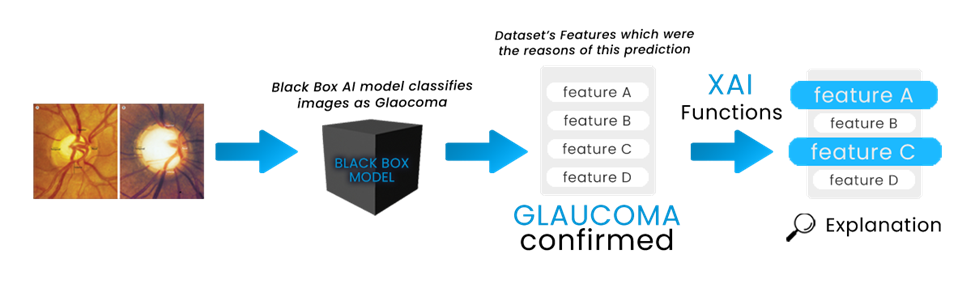}}
        \caption{Blackbox models decision making explanation through LIME}
        \label{BlackboxLime}
        \end{figure}

        We have employed Deep Learning or FCNNs in our work which is a BlackBox function. Generally, Black boxes work excellently but their structure won’t give you any insights that will explain how the function is being approximated. For this, we have used LIME which is one of the most popular XAI-based python libraries. There are a lot of XAI frameworks that explain the BlackBox model’s insights by features. XAI functions work well in terms of explaining complex classification models. In short, these functions generate an explanation through charts of graphs for a complex model's prediction which are also pretty fast. Figure \ref{BlackboxLime} represents how black boxes actually work with the help of LIME.

        Here we can see BlackBox models generate a result or output based on some features from the given/training datasets. And through lime, we can have a visualization from which features the output was based on. In our Glaucoma dataset, we have some features for Suspicious glaucoma and Non-glaucoma. In both sections, we have fundus images, and labels as 1 as the confirmed glaucoma case and 0 as the Non-glaucoma case. To apply XAI, we took Fully Connected Neural Networks (FCNNs) as a black box AI model to predict glaucoma with the help of the data. To compile all of these classifications and determine the average of these scores to one single output, we will use ReLU non-linear activation function in the convolutional layers and the Softmax activation function in the output layers. Below a short rendition is being given for the above Deep Learning models.

        \begin{itemize}
          \item \textbf{Convolutional Neural Network (CNN)}: In deep learning, a convolutional neural network (CNN, or ConvNet) is a class of deep neural networks, most commonly applied to analyze visual imagery. We will classify the image data through this model.
        
          \item \textbf{Fully Connected Neural Network (FCNNs)}: Fully connected neural networks (FCNNs) are a type of artificial neural network where the architecture is such that all the nodes, or neurons, are in one layer, are connected to the neurons in the next layer \cite{Murphy}. This model will also help us to predict and output.
        
          \item \textbf{ReLU}: A rectified linear activation unit, or ReLU for short, is a node or unit that implements this activation function. Often, networks with hidden layers that use the rectifier function are referred to as rectified networks. The computational cost of adding more ReLUs increases linearly as the size of the CNN grows.
        
          \item \textbf{Softmax}: Softmax is a mathematical function that transforms a vector of integers into a vector of probabilities, with the probability of each value proportional to the vector's relative scale. The softmax function is most commonly used as an activation function in a neural network model in applied machine learning. The network is set up to produce N values, one for each classification task class, and the softmax function is used to normalize the outputs, turning them from weighted sum values to probabilities that total to one. Each value in the output of the softmax function is interpreted as the probability of membership for each class. This will compile all the convolutional layers of the FCNNs into a single output.
        \end{itemize}

        According to our Dataset, we will divide the data chronologically into training and testing data to classify glaucoma.. And through Lime, a XAI function, we will explain these black boxes.

    Here in Figure \ref{SystemModel}, we have shown the whole process from dataset preprocessing to compiled output through Softmax activation function. And with XAI functions, we will explain the black boxes through visualization charts of the used core features which were the main reasons behind the prediction.

\subsection{Dataset Description}

    As our data are mostly direct fundus images from LAG-Dataset \cite{Khan}. CNN is being used in this thesis for image classification, as it is a type of model which processes data such as images. Also, it automatically understands low-to high-level patterns of image classification. which helps us to extract higher representations for the image content.


    The dataset contains 4250 images for training, 302 images for testing and 302 images for validation. All of these images have separated into two folders for glaucoma and non glaucoma. Depicted from Table \ref{lagDataset} the label for “glaucoma” is 1 and for “non-glaucoma” is 0.
    
    \begin{table}[!t]
        \begin{center}
        \caption{Dataset Description}
        \begin{tabular}{|c|c|c|}
            \hline
            Class               & Label & Fundus Image \\ \hline
            Suspicious Glaucoma & 1     & 1711         \\ \hline
            Non Glaucoma        & 0     & 3143         \\ \hline
        \end{tabular}%
        \label{lagDataset}
        \end{center}
    \end{table}

\subsection{Architecture}
    In deep learning, a convolutional neural network (CNN, or ConvNet) is a class of artificial neural networks, most commonly applied to analyze visual imagery. In this study, a Transfer Learning approach is proposed. The data set’s size and features provide a perfect environment for implementing a transfer learning approach, allowing a pre-trained CNN with all of its weights to be utilized to develop a new transfer learning model specialized to identifying Glaucoma with a high degree of accuracy. \par
    We used pre-trained models from Tensorflow Keras implementation and through Transfer Learning we trained only the layers we need to train. All the model's weights were trained from the ImageNet dataset. After downloading the pre-trained model we have made every trained layer into untrainable layers and deleted the top layers to reuse the model. Then we use a Flatten layer to flatten every pre-trained layer of the keras model's into one and we used 3 Dense neurons with 100 layers in each of it for VGG-16, VGG-19, InceptionV3 and ResNet50. For DenseNet121 we used 1024 layers for first, 512 layers for second and 256 layers for third neuron. We used the “ReLU” activation function for the Convolutional layers. We also performed batch normalization and a dropout with the rate of 0.5 in each model. For predictions, we used the A Dense neuron with 2 layers in it and “Softmax” for activation function. For the Gradient Descent, we used the Adam optimizer. 
    
    \begin{figure}[!t]
    \centerline{\includegraphics[scale=0.3]{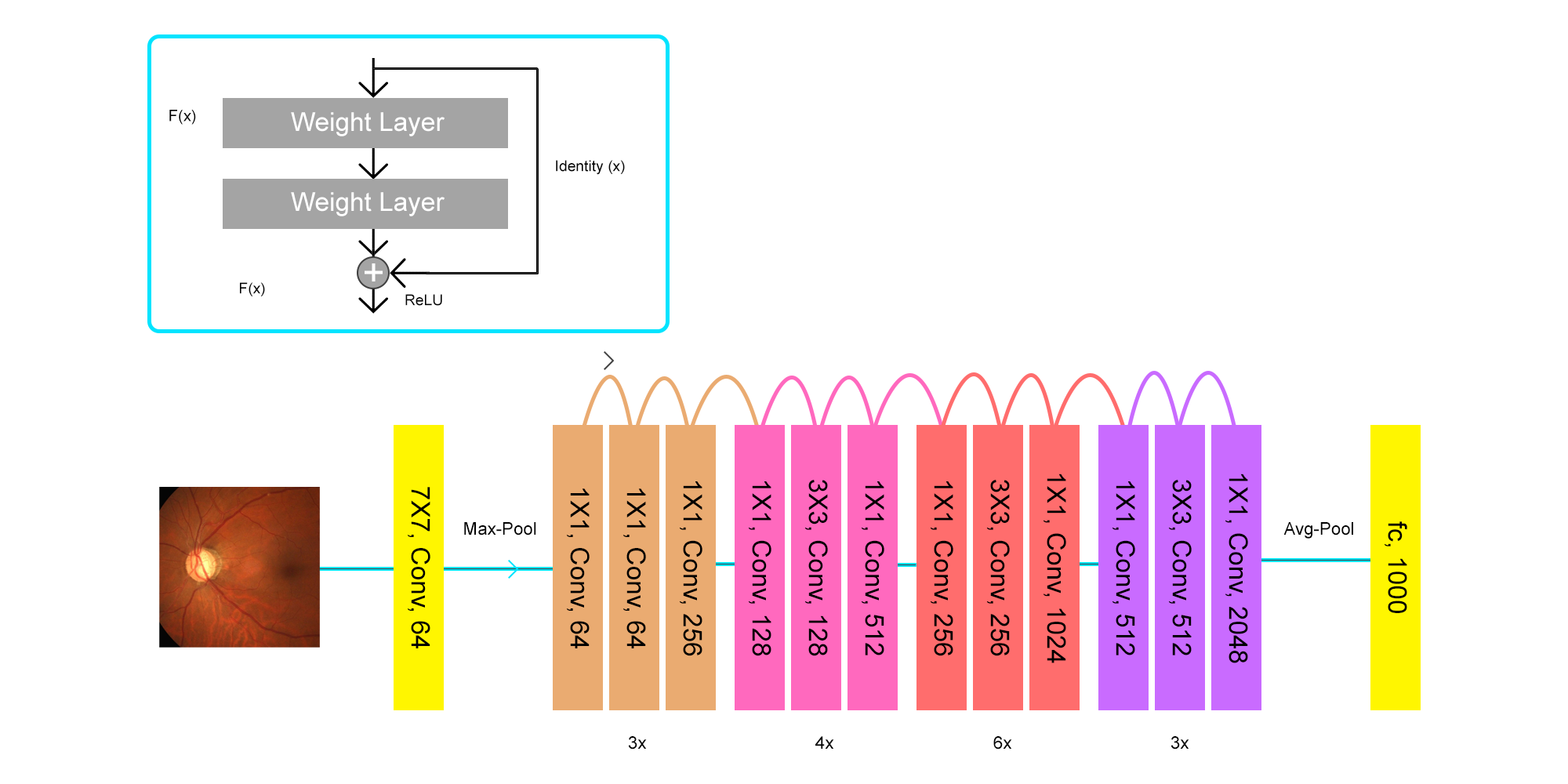}}
    \caption{Architecture of ResNet50 on Glaucoma}
    \label{ResNetArchitecture}
    \end{figure}

\section{Performance Evaluation}
    In the Figure \ref{ResNetArchitecture} we can see the architecture of RestNet50 which is our proposed model. However, we have utilized VGG-16, VGG-19, DenseNet121, InceptionV3 and ResNet50 models for our study which was compiled with Adam optimizer with the learning rate of 1e-5 in 50 epochs. After 50 epochs, RestNet50 managed to acquire the highest score among the other models with a validation accuracy of 94.7\%. Table \ref{ModelAccuracy} depicts the findings of our study. Table \ref{ModelPerformance} shows the misclassification count along with percentage among each of the model for both of the classes. Here, G = Glaucoma and n-G = Non-Glaucoma.
    
    \begin{table}[!t]
        \caption{MODEL ACCURACY AND LOSS}
        \begin{tabular}{|c|c|c|c|c|}
                \hline
                Model       & Accuracy & Train Accuracy & Train Loss & Valid Loss \\ \hline
                DenseNet121 & 86.81\%  & 88.83\%        & 31.18\%    & 24.10\%         \\ \hline
                InceptionV3 & 86.42\%  & 93.49\%        & 20.04\%    & 35.79\%         \\ \hline
                ResNet50    & 94.71\%  & 99.56\%        & 3.81\%     & 12.22\%         \\ \hline
                VGG-16      & 88.63\%  & 98.00\%        & 6.76\%     & 27.92\%         \\ \hline
                VGG-19      & 93.31\%  & 97.00\%        & 11.53\%    & 14.94\%         \\ \hline
        \end{tabular}
        \label{ModelAccuracy}
    \end{table}

    \begin{figure*}[!t]
        \centering
        \includegraphics[scale=1]{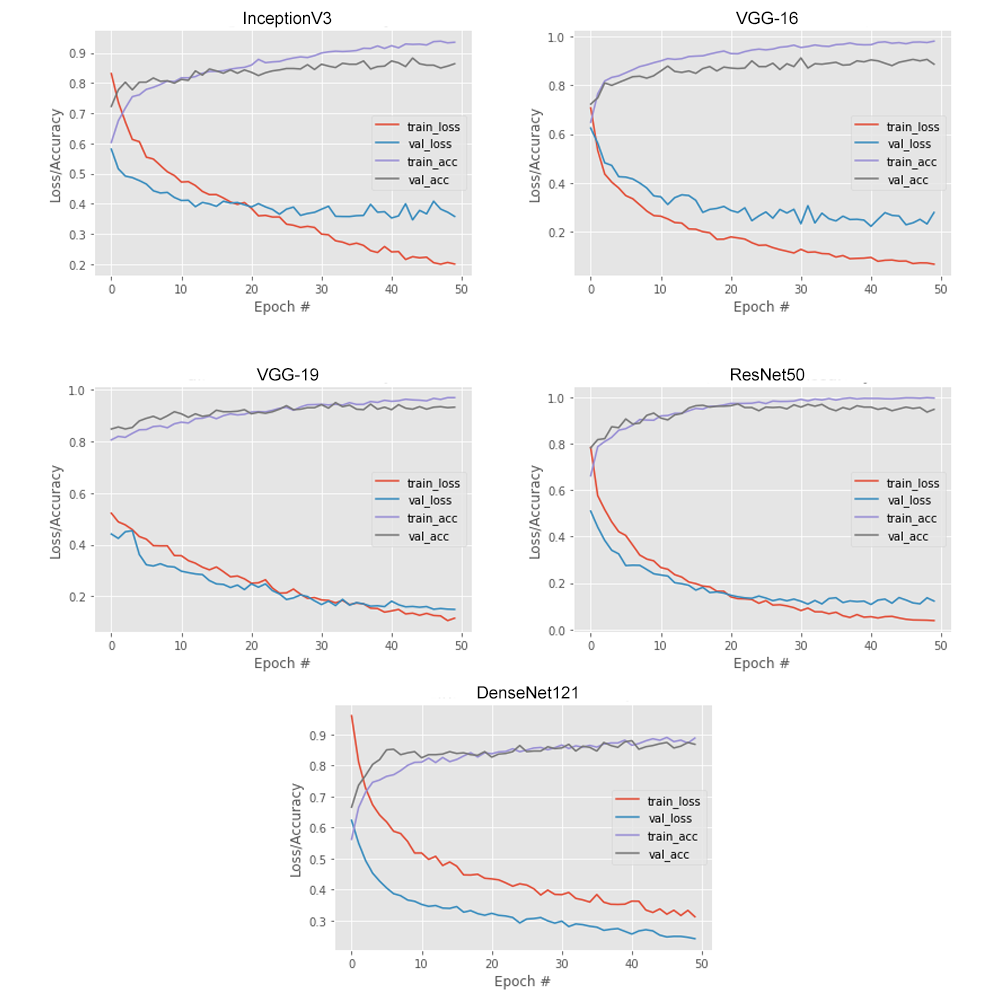}
        \caption{Transfer Learning Model's Accuracy and Loss Curve on Glaucoma Classification.}
        \label{TrainLossGraph}
    \end{figure*}
    
    Figure \ref{TrainLossGraph} represents the curve of accuracy, loss along with validation curve. Here, after extensive analysing of all the utilized model, we can state that VGG-19 and ResNet50 were the best-fitted models.

\begin{table}[!t]
    \begin{center}
    \caption{Model Performance with Misclassification}
    \begin{tabular}{ |c||c|c|c|c|} 
        \hline
         & Model & All  & G & n-G  \\
         &  & Misclassified &  Misclassified &  Misclassified \\
        \hline 
                & DenseNet121 & 9 (86.76\%)   & 4 (88.24\%)   & 5 (85.29\%)  \\
                & InceptionV3 & 151 (93.83\%) &  74 (93.95\%) & 77 (93.71\%) \\ 
        Test    & VGG-16      &  48 (97.65\%) &  22 (97.84\%) & 26 (97.45\%) \\
                & VGG-19      & 22 (94.12\%)  &  3 (91.18\%)  & 1 (97.06\%) \\
                & ResNet50    & 26 (95.59\%)  &  1 (97.06\%)  & 2 (94.12\%) \\
        \hline
        \hline
                    & DenseNet121 & 151 (93.83\%)  & 74 (93.95\%) & 77 (93.71\%) \\
                    & InceptionV3 & 48 (97.65\%) & 22 (97.84\%) & 26 (97.45\%) \\ 
        Valid         & VGG-16 & 47 (98.08\%)  & 20 (98.37\%) & 27 (97.79\%) \\ 
                    & VGG-19 & 17 (99.31\%)  & 14 (98.86\%) & 3 (33.75\%) \\ 
                    & ResNet50 & 0 (100\%)  & 0 (100\%) & 0 (100\%) \\         
        \hline 
    \end{tabular}
     \label{ModelPerformance}
    \end{center}
\end{table}


These are the single image predictions of all models - 





\section{XAI: Local Interpretable Model-Agnostic Explanations}

\noindent Now we will show the explanation for these preprocessed and misclassified images using an XAI \cite{Daglarli} framework, LIME. Then we will apply Lime again on a single predicted raw fundus -image directly from the test dataset (labelled) directory to see the difference between a correctly predicted fundus image \cite{Vejjanugraha} and wrong predicted fundus image. Given below are the misclassified image with preprocessing, Superpixels focused area and the model prediction explanation by Lime in DenseNet12.

\begin{figure*}[!t]
\centering
\includegraphics[scale=0.27]{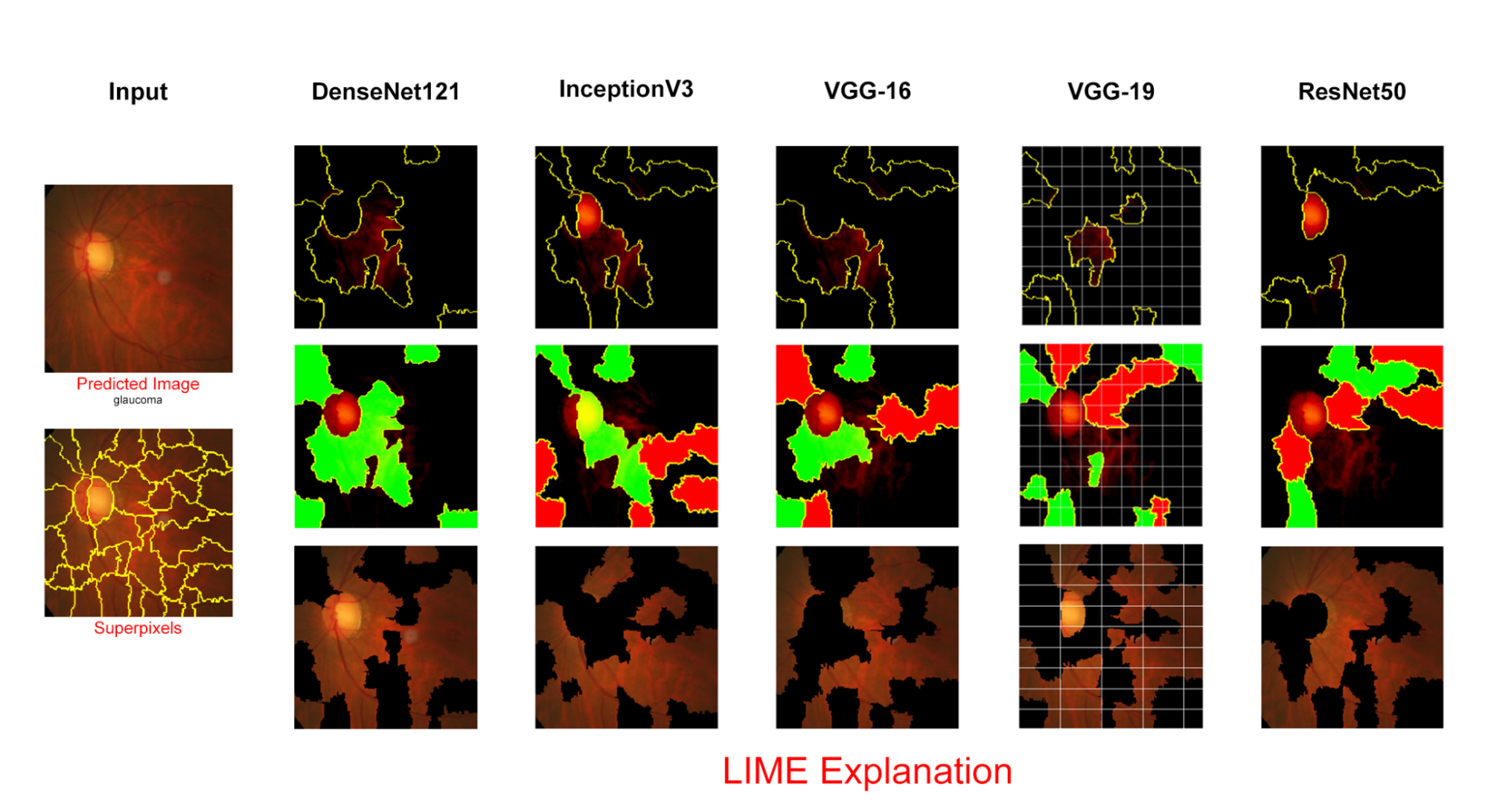}
\caption{Lime: Explanation for A Single Frame.}
\label{LIMEexplain}
\end{figure*}

\ref{LIMEexplain} depicts the LIME for a single image. Its goal is to make the predictions of machine learning models understandable to humans. The method can explain individual instances which makes it suitable for local explanations. LIME manipulates the input data and creates a series of artificial data containing only a part of the original attributes. Thus, in the case of text data, for example, different versions of the original text are created, in which a certain number of different, randomly selected words are removed. This new artificial data is then assigned to different categories (classified). Hence, through the presence or absence of certain keywords we can see their influence on the classification of the selected text. LIME gives the output as a list of explanations which reflects the contribution of each feature which resulted in the final prediction.



\section{Conclusion}
In this research, we have a proposed a model for detecting Glaucoma diseases based on transfer learning along  with a thorough comparison of the effectiveness of several pre-trained models for the classification of Glaucoma. In addition, we have employed Local Interpretable Model-agnostic Explanations(LIME) on all of the utilized model. This employment aids to build trust among the Medical professionals as they usually do not rely on deep learning based system as it tends to have black box characteristics. Thus its not possible to determine the key features behind a models prediction. However, LIME locates these crucial elements and creates a visual representation of their reasoning. Our proposed system is trained on a benchmark Glaucoma dataset and with ResNet50 we have managed to acquire a validation accuracy of 94.7\%. VGG-19 also managed to reach an validation accuracy of 93.3\%. We intend to advance this study in the future by enhancing the efficiency and ease of glaucoma detection. Till now we have detected Glaucoma and Non-Glaucoma at a satisfactory accuracy rate using multiple models but in near future we are planning to develop a web application which will help people to identify Glaucoma by just uploading the images. Moreover, we are planning to work on more datasets with improved accuracy. 


\end{document}